\documentclass[dvipsnames]{article} %
\usepackage{colm2024_conference}

\usepackage{booktabs}
\usepackage{graphicx}
\usepackage{enumitem}
\usepackage{wrapfig}
\usepackage{algorithm}
\usepackage{algpseudocode}

\usepackage{microtype}
\usepackage{amsmath}
\usepackage{colortbl}
\usepackage[utf8]{inputenc}
\definecolor{lightgray}{rgb}{0.9,0.9,0.9}
\usepackage{caption}
\usepackage[table]{xcolor}

\usepackage{subcaption}
\usepackage{setspace}
\usepackage{url}
\usepackage{multirow}
\usepackage{colortbl}
\usepackage{tabularx}
\usepackage{blindtext}
\usepackage{pgfplots}
\pgfplotsset{compat=1.18} 
\usepackage{tikz}
\usetikzlibrary{er,positioning,bayesnet}
\usepackage{makecell}
\usepackage{tipa}
\usepackage{siunitx}
\usepackage{nicefrac}
\usepackage{tocloft}
\usepackage{listings}
\usepackage[raster,skins]{tcolorbox} %
\usepackage{xltabular}
\usepackage{adjustbox}
\usepackage{xurl}
\usepackage{rotating}
\usepackage[normalem]{ulem}
\useunder{\uline}{\ul}{}

\usepackage{amsmath,amsfonts,bm}

\def\eqref#1{equation~\ref{#1}}
\def\1{\bm{1}}

\DeclareMathAlphabet{\mathsfit}{\encodingdefault}{\sfdefault}{m}{sl}
\SetMathAlphabet{\mathsfit}{bold}{\encodingdefault}{\sfdefault}{bx}{n}

\usepackage{float}
\usepackage{pifont}

\newcommand{\tabincell}[2]{\begin{tabular}{@{}#1@{}}#2\end{tabular}}

\newcommand*\justify{%
\fontdimen2\font=0.4em% interword space
\fontdimen3\font=0.2em% interword stretch
\fontdimen4\font=0.1em% interword shrink
\fontdimen7\font=0.1em% extra space
\hyphenchar\font=`\-% allowing hyphenation
}

\renewcommand{\texttt}[1]{%
\begingroup
\ttfamily
\begingroup\lccode`~=`/\lowercase{\endgroup\def~}{/\discretionary{}{}{}}%
\begingroup\lccode`~=`[\lowercase{\endgroup\def~}{[\discretionary{}{}{}}%
\begingroup\lccode`~=`.\lowercase{\endgroup\def~}{.\discretionary{}{}{}}%
\catcode`/=\active\catcode`[=\active\catcode`.=\active
\justify\scantokens{#1\noexpand}%
\endgroup
}

\newcommand{\hflink}{https://huggingface.co/sunshk/openpi_comet}
\newcommand{\hflinkdata}{https://huggingface.co/datasets/delinqu/comet-1.5k}
\newcommand{\ghlink}{https://github.com/mli0603/openpi-comet}

\usepackage{tablefootnote}

\title{Openpi Comet: Competition Solution For 2025 BEHAVIOR Challenge}

\author{
\bf Team Comet
}

\begin{document}

\maketitle

\begin{abstract}
    The 2025 BEHAVIOR Challenge is designed to rigorously track progress toward solving long-horizon tasks by physical agents in simulated environments.
    BEHAVIOR-1K focuses on everyday household tasks that people most want robots to assist with and these tasks introduce long-horizon mobile manipulation challenges in realistic settings, bridging the gap between current research and real-world, human-centric applications. 
    This report presents our solution to the 2025 BEHAVIOR Challenge in a very close 2nd place and substantially outperforms the rest of the submissions. 
    Building on $\pi_{0.5}$, we focus on systematically building our solution by studying the effects of training techniques and data. 
    Through careful ablation studies, we reveal the scaling benefits in both the pre-training and post-training phases, leading to a validation Q-score of 0.345, significantly surpassing previous state-of-the-art performance.
    We summarize our practical lessons and design recommendations that we hope will provide actionable insights for the broader embodied AI community when adapting powerful foundation models to complex embodied scenarios.
\end{abstract}
\section{Introduction}
\label{sec:intro}

Vision-Language-Action (VLA) models~\citep{brohan2022rt,brohan2023rt,team2024octo,kim2024openvla,qu2025spatialvla,black2024pi_0} have recently emerged as a unifying paradigm for robotic policy learning, leveraging large-scale robot datasets to acquire robust and generalizable manipulation and navigation capabilities. By integrating perception, language understanding, and control within a single end-to-end framework, VLAs bypass the need for hand-engineered modules and have demonstrated strong performance across a variety of embodied AI benchmarks. Despite this progress, most existing VLA systems are primarily optimized for short-horizon tasks, and their ability to scale to complex, temporally extended activities remains limited.

Long-horizon manipulation~\citep{zhao2025cot,zawalski2024robotic} introduces additional difficulties that fundamentally challenge current VLA designs. Such tasks require orchestrated sequences of interdependent behaviors, where compounding errors and shifting state distributions can degrade performance over time. A common approach is to decompose tasks into subtasks~\citep{lin2022diffskill,shi2023robocook,tie2025manual2skill} and train separate local policies. However, this strategy does not resolve the skill chaining problem~\citep{chen2024scar,konidaris2009skill}, which involves modeling and executing reliable transitions between subtasks while mitigating error accumulation. In addition, many solutions proposed for skill chaining rely on online adaptation or modular architectures, and these methods are often incompatible with the large-scale, offline, end-to-end training paradigm that underpins modern VLA models. Consequently, achieving reliable long-horizon performance while preserving scalability and generality remains an open challenge.

The BEHAVIOR Challenge, built upon the BEHAVIOR-1K~\citep{li2024behavior} benchmark, provides an rigorous benchmark for this problem. It features realistic household environments containing complex object interactions, and evaluates agents on 50 long-horizon tasks that reflect human-centered daily activities. Each task requires multi-step reasoning, precise manipulation, and coordinated navigation, making success highly dependent on robust long-horizon policy execution. With 10,000 expert demonstrations and a standardized evaluation protocol, the challenge places strong emphasis on generalization, control robustness, and error tolerance. These capabilities remain difficult for current VLA models to achieve consistently.

In this report, we examine how far a strong publicly available VLA backbone can be pushed on long-horizon tasks using careful data, training, and inference design within a simple end-to-end training pipeline. We treat the BEHAVIOR Challenge as a case study in adapting powerful but generic foundation policies to a complex embodied benchmark. Through systematic exploration of training configurations, pre-training choices, and inference strategies, we show that our solution completes 22 tasks out of the 50 household tasks, achieving a Q-score of 0.2514 in the competition \autoref{tab:competition_result}. After the challenge, we have refined our post-training algorithm and achieved a \textbf{validation Q-score of 0.3453}, substantially outperforming all previous results.
\section{Architecture}
\label{sec:method}

\begin{figure}[ht]
    \centering
    \includegraphics[width=\linewidth]{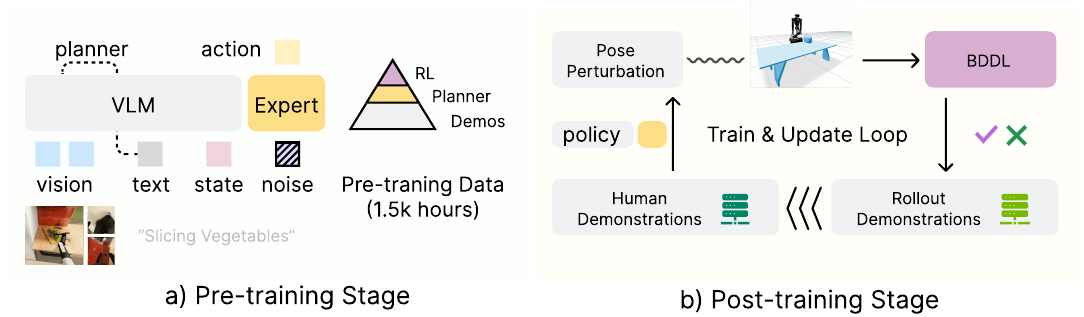}
    \caption{(a) Pre-training on large-scale heterogeneous data, including 1.1K 
    hours of human demonstrations and $\sim$0.4K hours of additional planner and offline RL trajectories. (b) RFT post-training: perturb initial poses, roll out the policy, and retain successful episodes to iteratively augment the dataset.}
    \label{fig:pipeline}
\end{figure}

As shown in~\autoref{fig:pipeline}(a), we adopt the $\pi_{0.5}$ as the base policy of our system. $\pi_{0.5}$ follows the standard VLA design paradigm, combining a visual encoder for multi-view robot observations with a language encoder for task instructions, and fusing these modalities into a shared representation that conditions the action expert. The action expert is implemented as a transformer-style network that ingests features and denoises the low-level continuous control actions at each timestep. This end-to-end architecture allows perception, language understanding, and control to be trained jointly from large-scale robot datasets, and $\pi_{0.5}$ further enhances generalization by being pretrained on heterogeneous data spanning multiple embodiments, environments, and tasks. The dataset includes the \textbf{1k hours human demonstrations officially provided by the BEHAVIOR Challenge}, as well as \textbf{our additional motion-planner trajectories} and \textbf{offline RL rollouts}, which together supply rich long-horizon behaviors and diverse manipulation strategies crucial for robust policy learning.

For post-training, we adopt an iterative RFT procedure as illustrated in \autoref{fig:pipeline}(b). Starting from the official human demonstrations, we introduce random pose perturbations and roll out the pretrained policy under these disturbed initial conditions. Successful episodes are retained as additional training data, progressively forming an offline data flywheel that continually improves the robustness and coverage of the policy.

\begin{table}[tb]
    \centering
    \caption{Results of 2025 BEHAVIOR Challenge for standard track and our post-challenge solution. Q-score for the test set is used for final ranking. Best viewd in color.}
    \begin{tabular}{cccccc}
        \toprule
        \multirow{2}{*}{\textbf{Rank}} & \multirow{2}{*}{\textbf{Team}} & \multicolumn{2}{c}{\textbf{Full Task Success Rate}} & \multicolumn{2}{c}{\textbf{Q-Score}} \\
                                       &                                & Validation                                          & Test                                 & Validation & \textcolor{blue}{\textbf{Test}}       \\
        \midrule
                                      & \textcolor{olive!70!black}{Comet (ours, post-challenge)}               &  \textcolor{olive!70!black}{0.1500} &                               & \textcolor{olive!70!black}{0.3453} %\tablefootnote{Building upon our submission, we further refined the training strategy, leading to a significantly higher Q-score of 0.3453.} 
                                      &  \\ \midrule
        1                              & Robot Learning Collective      & 0.1120                                              & 0.1240                               & 0.2605     & 0.2599     \\
        \textcolor{olive!70!black}{2}                              & \textcolor{olive!70!black}{Comet (ours)}               & \textcolor{olive!70!black}{0.1440}                                              & \textcolor{olive!70!black}{0.1140}                               & \textcolor{olive!70!black}{0.1830}\tablefootnote{Due to limited time by the submission deadline, we could not finish evaluation across all tasks.}     & \textcolor{olive!70!black}{0.2514} \\
        3                              & SimpleAI Robot                 & 0.1400                                              & 0.1080                               & 0.1943     & 0.1591     \\
        4                              & The North Star                 & 0.1280                                              & 0.0760                               & 0.1702     & 0.1204     \\
        \bottomrule
    \end{tabular}
    \label{tab:competition_result}
\end{table}

\section{Dataset}
\label{sec:dataset}
\begin{figure}[ht]
    \vspace{-2ex}
    \begin{minipage}[c]{0.32\linewidth}
        \centering
        \includegraphics[width=\linewidth]{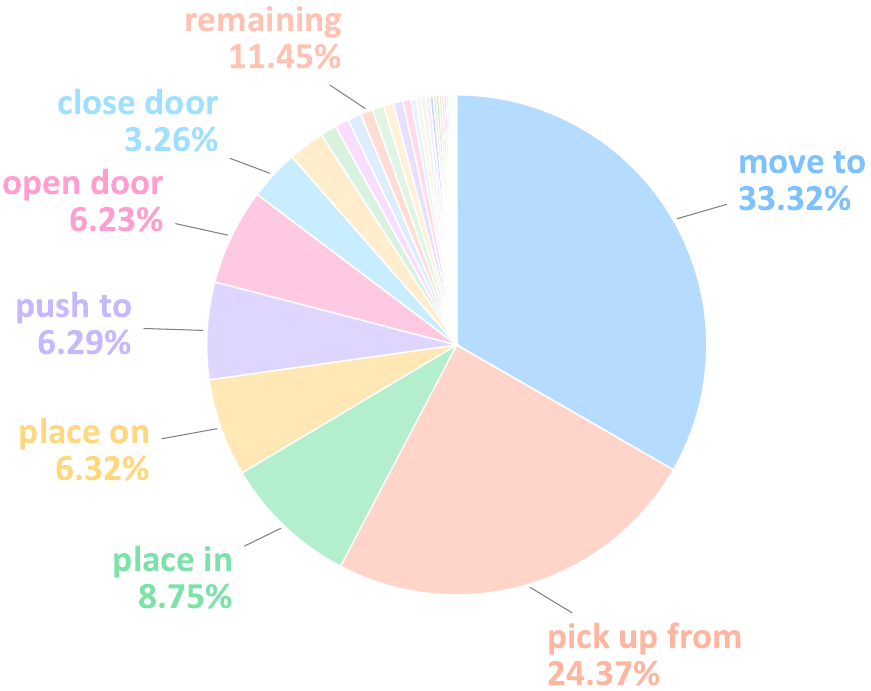}
        \caption*{(a) Skill distribution.}
    \end{minipage}
    \hfill
    \begin{minipage}[c]{0.67\linewidth}
        \centering
        \includegraphics[width=\linewidth]{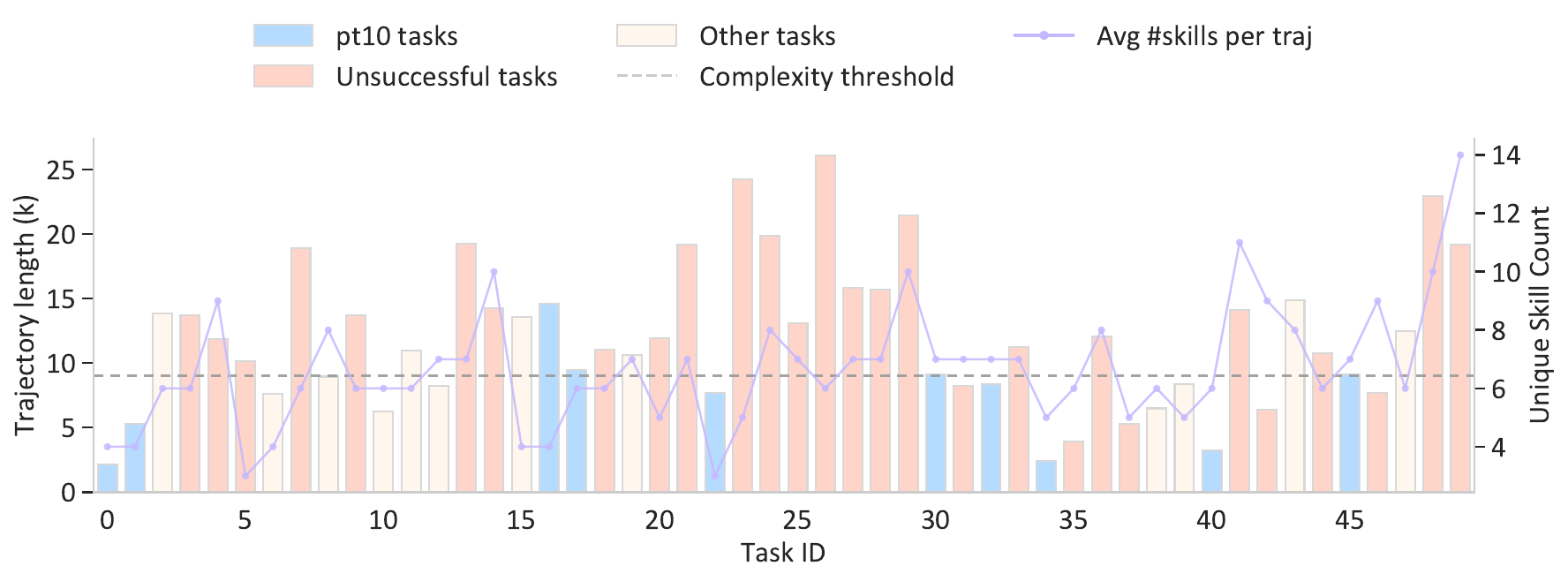}
        \caption*{(b) Dataset statistic.}
    \end{minipage}

    \caption{Dataset statistics and skill distribution for BEHAVIOR Challenge @ NeurIPS 2025. (a) Proportion of video frames occupied by each skill across the entire dataset. (b) Per-task distribution, showing the average trajectory length (in frames) and the average number of unique skills per trajectory.}
    \label{fig:dataset}
\end{figure}

The BEHAVIOR-1K benchmark provides 1,000 realistic household activities instantiated in 50 fully interactive 3D scenes with over 10k objects, designed to stress long-horizon mobile manipulation and high-level reasoning in human-centric environments. The NeurIPS 2025 BEHAVIOR Challenge selects 50 representative tasks from BEHAVIOR-1K and supplies 10,000 teleoperated expert demonstrations (200 per task, over 1,200 hours) with multi-modal observations and fine-grained skill annotations. Policies are evaluated in simulation by the task success rate defined via BDDL goal predicates, emphasizing completion of entire activities rather than short-horizon subroutines.

Beyond the official teleoperated dataset, we further incorporate $\sim$3.6K trajectories composed of motion-planner demonstrations and offline RL rollouts. The planner data provides precise low-noise manipulation sequences, while the RL rollouts introduce broader behavioral variability. Together, these additional trajectories substantially enrich the state–action coverage beyond human demonstrations.

To better understand the learning landscape, we analyze the demonstration dataset at the level of semantic skills and per-task complexity. As shown in \autoref{fig:dataset}(a), the distribution is highly imbalanced: \texttt{move to} and \texttt{pick up from} dominate with roughly 33.3\% and 24.4\% of frames, followed by \texttt{place in} (8.8\%) and a long tail of infrequent skills (11.5\%). \autoref{fig:dataset}(b) reports per-task statistics: many tasks require trajectories of several hundred frames and typically involve 5–10 distinct skills, with some exceeding 12. We treat tasks with an average length below 250 frames as relatively simple, since they involve shorter horizons and fewer skill compositions, and they are particularly useful for quickly bringing up our system and validating basic policy behavior in the simulator. In contrast, several tasks, including Task 48 and Task 49, fall into a clearly harder regime, characterized by long-horizon execution and rich mixtures of navigation and manipulation skills (e.g., moving between rooms, opening/closing containers, and rearranging multiple objects). These high-complexity tasks provide a stringent testbed for evaluating policies’ ability to handle extended temporal credit assignment and frequent skill switching.
\section{Experiment}
\label{sec:experiment}
\begin{minipage}[c]{0.48\textwidth}
    In this section, we present details of our implementations~(\autoref{sec:implementation}), pre-training~(\autoref{ssec:pretrain}) and post-training~(\autoref{ssec:posttrain}). We report the results across different stages in~\autoref{tab:validation_qscore} and present detailed ablations in~\autoref{ssec:ablation}.
\end{minipage}
\hfill
\begin{minipage}[c]{0.48\textwidth}
    \centering
    \captionof{table}{Validation Q-scores across training stages.}
    \begin{tabular}{cccc}
        \toprule
        \textbf{Pre-training} & \textbf{Post-training} & \textbf{Theoretical best}\\
        \midrule
        0.192                 & 0.345                    & 0.611\\
        \bottomrule
    \end{tabular}
    \label{tab:validation_qscore}
\end{minipage}

\subsection{Implementation details}
\label{sec:implementation}
We adopt $\pi_{0.5}$ implemented in JAX as our policy, and all experiments are conducted on NVIDIA H200 GPUs with a per-device batch size of 64, using cosine decay scheduler. For single task SFT, we train for 15k–20k steps. For all multi-task pre-training, we use 50k training steps, with a learning rate of $2.5\times10^{-5}$. During both SFT and the subsequent RFT adaptation, we use a reduced learning rate of $2.5\times10^{-6}$ on 8 GPUs to ensure stable fine-tuning. 

To accelerate offline rollout generation, we parallelize evaluation across the 10 test instances by distributing them over multiple GPUs, which mitigates the extremely slow simulation rate of BEHAVIOR—where evaluating a single task can otherwise take from one hour to nearly a full day, making online RL potentially inefficient under this simulator~\cite{yu2025rlinf}.

\subsection{Pre-training}
\label{ssec:pretrain}

To study how pre-training task coverage and diversity affect long-horizon performance, we compare single-task pre-training \texttt{pt1} with three multi-task pre-training settings on the BEHAVIOR Challenge: \texttt{\#pt7}, \texttt{\#pt10}, and \texttt{\#pt50}. Each setting trains a single VLA policy on demonstrations from 7, 10, or all 50 challenge tasks, For each configuration, we train the model end-to-end on the corresponding task subset and then evaluate on the same 50-task challenge protocol. This design allows us to isolate how increasing the number and heterogeneity of pre-training tasks influences task success, while keeping the overall training recipe fixed.

\begin{minipage}[c]{0.48\textwidth}
    As shown in~\autoref{fig:pt-exp}, \texttt{pt1} trains the policy only on demonstrations from a single target task; this single-task finetuning regime yields the lowest average success and only produces successful rollouts on 2 tasks, indicating that purely task-specific adaptation is insufficient for robust long-horizon control. Building on this baseline, \texttt{pt7} pre-trains on a small subset of relatively short-horizon BEHAVIOR Challenge tasks such as \texttt{bringing water}, \texttt{cook hot dogs}, and \texttt{make microwave popcorn}, which share similar interactions with common household objects. As the pre-training set expands from \texttt{pt7} to \texttt{pt10} and finally \texttt{pt50}, more tasks begin to exhibit successful rollouts, showing that broader task coverage improves the model’s ability to generalize. In particular, \texttt{pt10} augments \texttt{pt7} with additional, slightly more complex tasks (e.g., \texttt{moving boxes to storage}, \texttt{hanging pictures}), while \texttt{pt50} uses demonstrations from all 50 challenge tasks, including rare and highly compositional activities such as \texttt{rearranging kitchen furniture} and \texttt{setting the fire}, thus exposing the policy to the full long-horizon distribution during pre-training.
\end{minipage}
\hfill
\begin{minipage}[c]{0.48\textwidth}
    \centering
    \includegraphics[width=\linewidth]{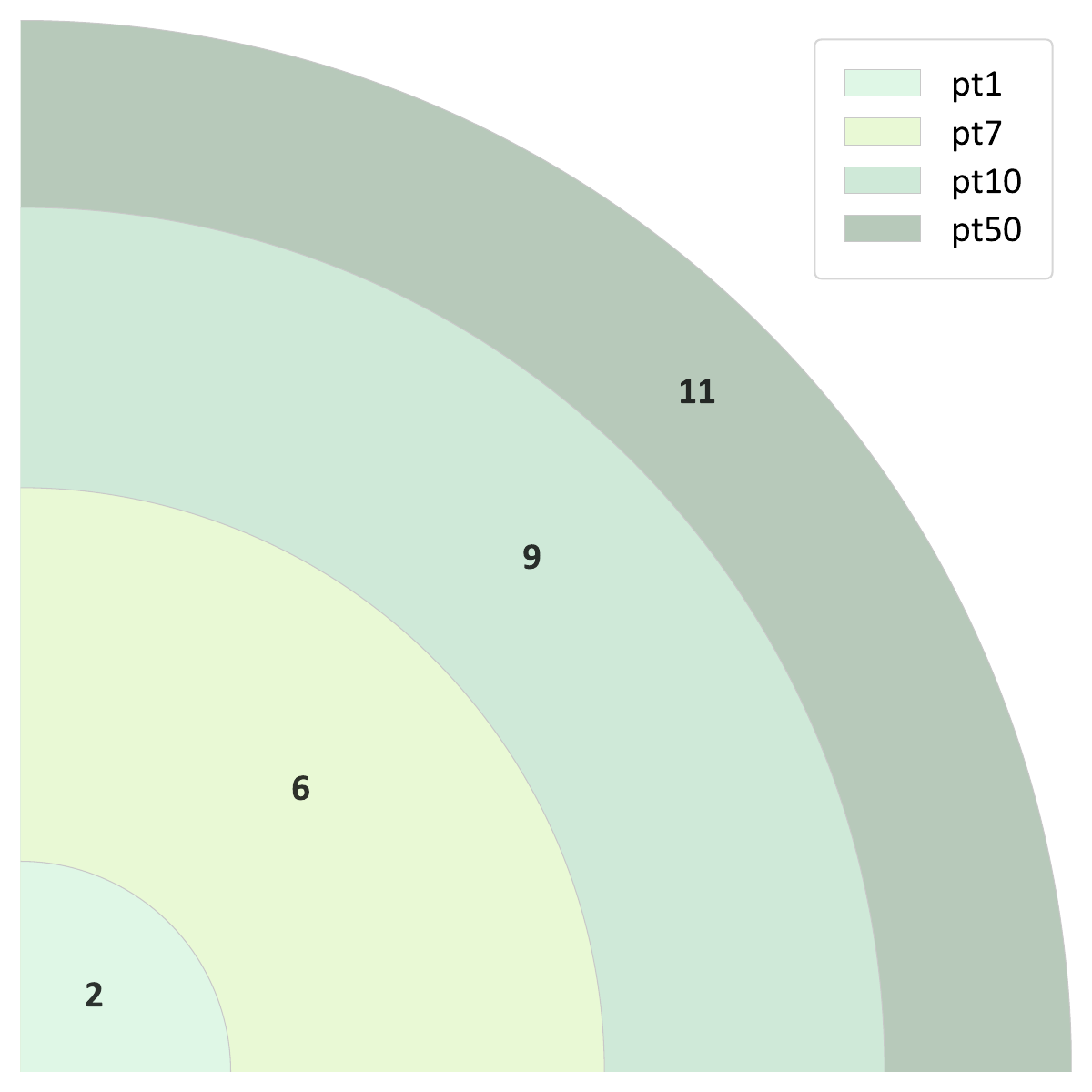}
    \captionof{figure}{Training subset coverage and the number of successful tasks.}
    \label{fig:pt-exp}
\end{minipage}

After pre-training, we reach a validation Q-score of 0.192 on the validation set, as shown in \autoref{tab:validation_qscore}.

\subsection{Post-training}
\label{ssec:posttrain}

\begin{minipage}[c]{0.48\textwidth}
    Despite the recent progress in online RL~\citep{chen2025pi_}, its low sample efficiency makes it impractical for the BEHAVIOR challenge. Moreover, online RL requires a heterogeneous compute setup: GPUs with RT Cores are needed for simulation, while GPUs with Tensor Cores are needed for model training and rollouts. Given these constraints, we instead adopt rejection sampling fine-tuning (RFT), a technique shown to be effective in both LLMs and VLMs~\citep{ahn2024large, touvron2023llama, azzolini2025cosmos}. An overview of our RFT setup is provided in \autoref{alg:rft}. Starting from all the provided train and validation human demonstrations for each scene, we randomly perturb the robot’s initial pose and use our pre-trained policy to perform rollouts under these perturbed configurations. Using both train and validation set helps avoid overfitting to the validation set and provides better signals of the model quality. Successful rollouts are retained as additional demonstrations. We perform $N=3$ rounds of RFT in total, collecting on average $T=8500$ trajectories per round, and eventually selected 1469 trajectories for training after de-duplication and task balancing.
\end{minipage}
\hfill
% --- Right Side: Algorithm with Borders ---
\begin{minipage}[c]{0.46\textwidth}
    \vspace{0pt} % Aligns top of algorithm with top of image
    % The [H] prevents floating, allowing it inside minipage while keeping borders
    \begin{algorithm}[H]
        \caption{The RFT Algorithm}
        \label{alg:rft}
        \begin{algorithmic}[1]
            \State Initialize $\mathcal{D} \leftarrow $ human demos.
            \State Initialize $\pi_1$ to pre-trained $\pi_{pt}$.
            \For{$i = 1$ to $N$}
            \State $\mathcal{D}_{i} \leftarrow \emptyset$
            \For{$t = 1$ to $T$}
            \State Sample initial state $s_0$ from $\mathcal{D}$.
            \State $s'_0 \leftarrow s_0 + \epsilon$.
            \State Rollout $\tau$ from $s'_0$ using $\pi_i$.
            \State $\mathcal{D}_{i} \leftarrow \mathcal{D}_{i} \cup \mathcal{D}_{\tau}$ if successful.
            \EndFor
            \State $\mathcal{D} \leftarrow \mathcal{D} \cup \mathcal{D}_{i}$.
            \State Train $\pi_{i+1}$ on $\mathcal{D}$.
            \EndFor
            \State \Return best $\pi_i$ on validation.
        \end{algorithmic}
    \end{algorithm}
\end{minipage}

\begin{figure}[p]
    \centering
    \includegraphics[width=\linewidth]{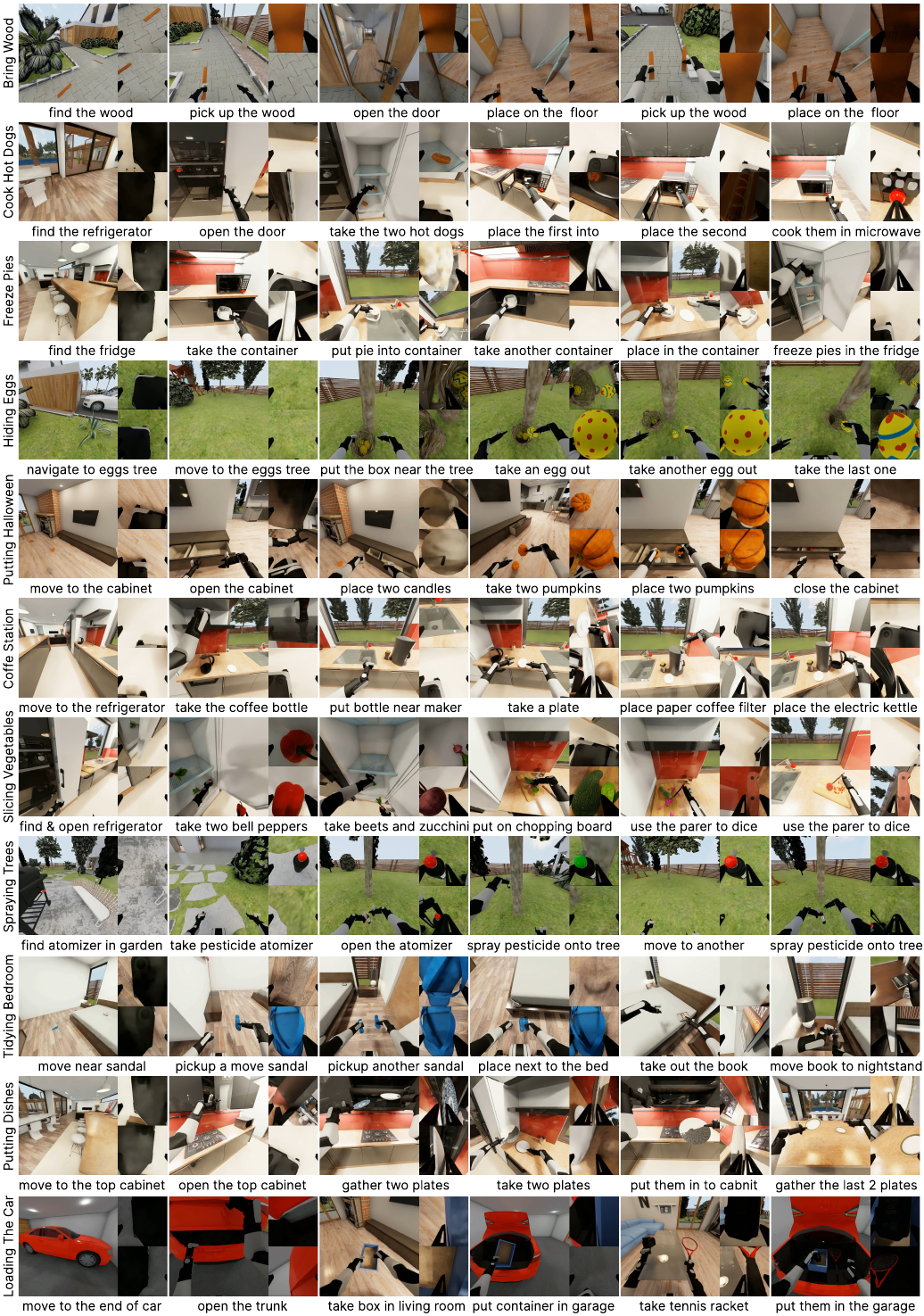}
    \caption{Rollout Examples on BEHAVIOR-1K. Our policy successfully handles long-horizon, multi-stage household tasks involving navigation, fine manipulation, and tool use, demonstrating robust execution across diverse household activities.}
    \label{fig:rollout_examples}
\end{figure}

During the challenge, our suite of models achieves a validation Q-score of 0.224 after post-training. After the challenge, we further refine the task balancing strategy and achieve a significantly higher \textbf{Q-score of 0.345 on the validation set}, using only two model checkpoints. Representative policy execution visualizations are provided in~\autoref{fig:rollout_examples}. Beyond improving overall performance, Rejection Sampling Fine-Tuning (RFT) also serves as a diagnostic mechanism by aggregating successful instances across historical checkpoints. As shown in~\autoref{tab:validation_qscore}, the theoretical best results we could obtain is a validation Q-score of 0.611 and a success rate of 0.35, revealing substantial headroom for further optimization. Further scaling up the model size could be helpful to realize the full potential of our post-training strategy.

\begin{figure}[t]
    \vspace{-4ex}
    \centering
    \includegraphics[width=\linewidth]{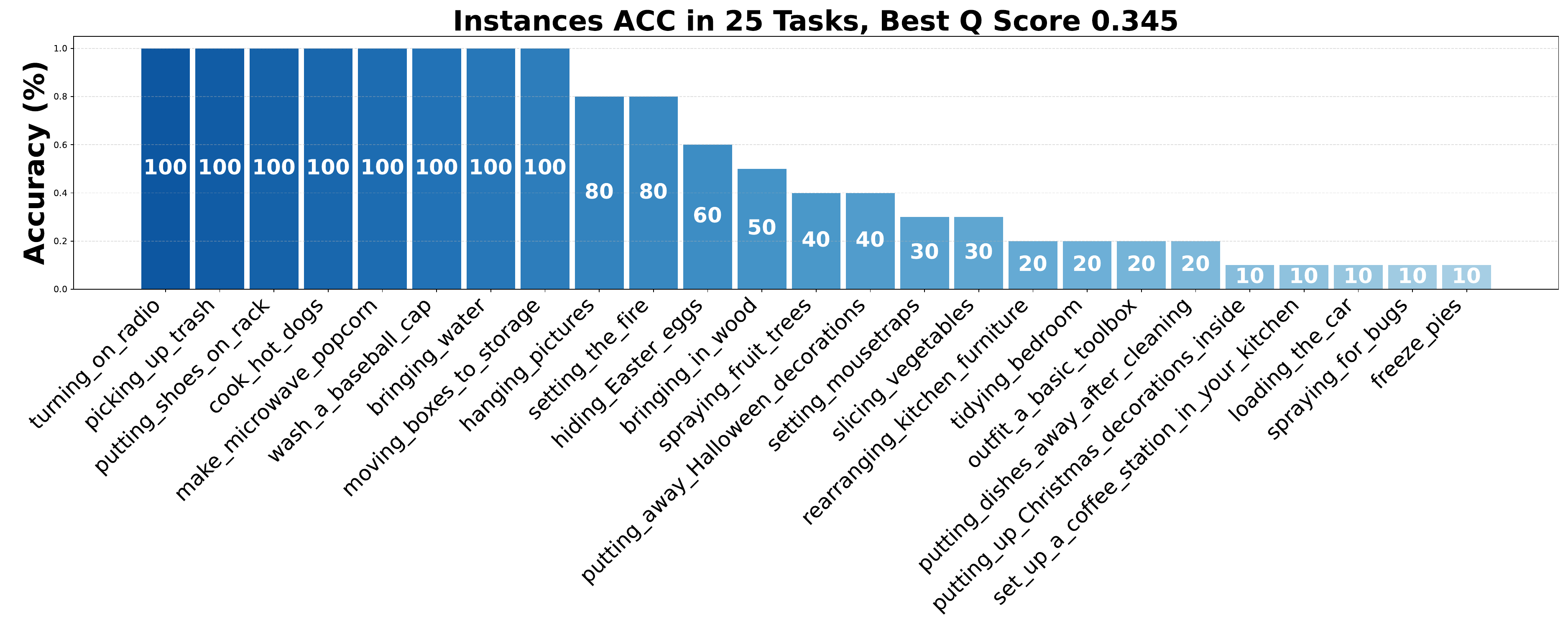}
    \caption{We can achieve an aggregated validation Q-score of 0.345 and a validation success rate of 15\% across 50 tasks. Our models can perform easy tasks such as \texttt{turnining\_on\_radio} robustly. Zero-success rate tasks are not shown. %, serving as the upper bound of our model's performance.
    }
    \label{fig:rollout_acc}
\end{figure}

\subsection{Ablations}
\label{ssec:ablation}
Beyond pre-training and post-training, we find that low-level design decisions in training and inference have a substantial impact on the performance. \autoref{tab:ablation} reports a set of controlled ablations along four such axes: Action Horizon and Input Modality for the training procedure, and Control Mode and Image Resolution for the inference strategy.

\textbf{Control mode (\#1)}. Temporal Ensemble and Receding Temporal fail to produce stable closed-loop behavior and result in near-zero success rates. In contrast, the Receding Horizon scheme, which executes all the predicted action segment and performs re-planning after finishing manipulation, significantly improves performance. This result highlights the necessity of continuous feedback for long-horizon manipulation and shows that smoothing or averaging open-loop predictions quickly accumulates error.

\textbf{Action horizon (\#2)}. Varying the action horizon reveals a non-monotonic relationship between prediction length and downstream control performance. Moderate horizons strengthen long-horizon manipulation by allowing the model to anticipate multi-stage behaviors, whereas excessively long horizons introduce conflicting temporal dependencies that compromise stable control. These findings indicate that the horizon length must be chosen to balance future awareness with the reliability of short-term control. We empirically find that setting the receding horizon to 32 gives us the best results.

\begin{table}[tb]
    \caption{Ablation study of control mode, action horizon, input modality, and image resolution on the \texttt{turning on radio} task.}
    \centering
    \setlength\tabcolsep{3pt}
    \begin{tabular}{@{}lcccc|c@{}}
        \toprule
        Settings             & Control Mode                      & Action Horizon      & Input Modality       & Image Resolution                       & Success Rate \\
        \midrule
        \multirow{3}{*}{\#1} & Temporal Ensemble                 & \multirow{3}{*}{50} & \multirow{3}{*}{RGB} & \multirow{3}{*}{\tabincell{c}{Head:224                \\Wrist:224}} & 0.00                 \\
                             & Receding Temporal                 &                     &                      &                                        & 0.00         \\
                             & Receding Horizon                  &                     &                      &                                        & \bf 0.25     \\
        \midrule
        \multirow{4}{*}{\#2} & \multirow{4}{*}{Receding Horizon} & 8                   & \multirow{4}{*}{RGB} & \multirow{4}{*}{\tabincell{c}{Head:224                \\Wrist:224}} & 0.00  \\
                             &                                   & 16                  &                      &                                        & 0.10         \\
                             &                                   & 50                  &                      &                                        & 0.25         \\
                             &                                   & 32                  &                      &                                        & \bf 0.30     \\
        \midrule
        \multirow{3}{*}{\#3} & \multirow{3}{*}{Receding Horizon} & \multirow{3}{*}{32} & RGB+Depth Image      & \multirow{3}{*}{\tabincell{c}{Head:224                \\Wrist:224}}                          & 0.20   \\
                             &                                   &                     & RGB+Point Cloud      &                                        & 0.30         \\
                             &                                   &                     & RGB                  &                                        & \bf 0.30     \\
        \midrule
        \multirow{4}{*}{\#4} & \multirow{4}{*}{Receding Horizon} & \multirow{4}{*}{32} & \multirow{4}{*}{RGB} & \multirow{2}{*}{\tabincell{c}{Head:224                \\Wrist:224}}                          & \multirow{2}{*}{0.30}   \\
                             &                                   &                     &                      &                                        &              \\
                             &                                   &                     &                      & \multirow{2}{*}{\tabincell{c}{Head:720                \\Wrist:480}}                          & \bf \multirow{2}{*}{0.60}   \\
                             &                                   &                     &                      &                                        &              \\
        \bottomrule
    \end{tabular}
    \label{tab:ablation}
\end{table}

\textbf{Input modality (\#3)}. We find that reconstructed point clouds improve performance compared to depth map as an input modality, which suggests that explicit geometric structure provides more informative cues for object-centric manipulation. However, the improvement over RGB-only inputs is limited, while the computational and latency overhead is considerable. Overall, explicit 3D geometry offers benefits in cluttered scenes, but the gains are not consistently large enough to justify the increased system complexity.

\textbf{Image resolution (\#4)}. Increasing the spatial resolution of both head and wrist views leads to substantial gains in performance. Low-resolution inputs of 224 by 224 pixels provide only coarse visual information, whereas high-resolution inputs more than double the success rate. These results indicate that precise visual cues are critical for reliable manipulation in visually complex environments, and that high-fidelity perception plays an essential role in long-horizon tasks.

\textbf{Data process (\#5)}. We further evaluate several data-processing strategies, including action representation, action subsampling, removal of proprioceptive state inputs, and skill weighting about Manipulation and Navigation in~\autoref{tab:ablation_training}. We use default training setting as the folling: Receding Horizon, 32 action horizon, RGB input, and 224x224 image resolution. Results shows that relative-action parameterization and the removal of state both lead to worse results, suggesting that absolute action anchoring and explicit system observability are essential for stable closed-loop optimization. Action subsampling at 15 Hz accelerates execution but reduces temporal precision, leading to degraded manipulation accuracy. Finally, although reweighting the dataset toward manipulation segments shifts the empirical distribution, it yields no measurable improvement, indicating that simple resampling is insufficient to compensate for the structural difficulty and heterogeneous dynamics of long-horizon manipulation.

\begin{table}[tb]
    \caption{Ablation study of action representation, sampling frequency, proprioceptive state inputs, and skill weighting under default training settings in \autoref{tab:ablation}.}
    \centering
    \setlength\tabcolsep{3pt}
    \begin{tabular}{@{}lcccc|c@{}}
        \toprule
        Settings               & Action Representation           & Action Sampling        & State Input                 & Skill weighting               & Success Rate \\
        \midrule
        \multirow{2}{*}{\#5.1} & Delta Joint                     & \multirow{2}{*}{30 Hz} & \multirow{2}{*}{\checkmark} & \multirow{2}{*}{No Weighting} & 0.00         \\
                               & Absolute Joint                  &                        &                             &                               & \bf 0.30     \\
        \midrule
        \multirow{2}{*}{\#5.2} & \multirow{2}{*}{Absolute Joint} & 15 Hz                  & \multirow{2}{*}{\checkmark} & \multirow{2}{*}{No Weighting} & 0.00         \\
                               &                                 & 30 Hz                  &                             &                               & \bf 0.30     \\
        \midrule
        \multirow{2}{*}{\#5.3} & \multirow{2}{*}{Absolute Joint} & \multirow{2}{*}{30 Hz} & \ding{55}                   & \multirow{2}{*}{No Weighting} & 0.00         \\
                               &                                 &                        & \checkmark                  &                               & \bf 0.30     \\
        \midrule
        \multirow{2}{*}{\#5.4} & \multirow{2}{*}{Absolute Joint} & \multirow{2}{*}{30 Hz} & \multirow{2}{*}{\checkmark} & No Weighting                  & \bf 0.30     \\
                               &                                 &                        &                             & Manip:Nav=2:1                 & \bf 0.30     \\
        \bottomrule
    \end{tabular}
    \label{tab:ablation_training}
\end{table}
\section{Conclusion}
\label{sec:conclusion}

In this report, we presented our solution to the 2025 BEHAVIOR Challenge, adapting the publicly available $\pi_{0.5}$ backbone to a demanding long-horizon household benchmark and systematically studying how pre-training task coverage, post-training, and inference-time design choices affect performance on all tasks. Our experiments reveal that scaling pre-training over more numerous and diverse BEHAVIOR tasks significantly generates and unlocks success on rare, compositional activities. Additionally, We show that RFT as a post-training technique avoids the infrastructure issues from online RL yet yield trackable targets as well as significant performance boost.

Despite the promising results, we acknowledge that our Q-score is still far from perfect. We find that despite the effectiveness of RFT, the sampling efficiency is still too low. Paradigms such as DAgger or on-policy distillation where an expert policy potentially using privileged data could greatly improve the sampling efficiency. In addition, RL approaches that provide both positive and negative rewards could balance the learning.

We believe that combining strong VLA backbones with more structured long-horizon reasoning, richer post-training objectives, and better curriculum design will further close the gap between synthetic benchmarks and real-world deployment, and we hope our empirical findings provide practical guidance for scaling foundation policies to complex, human-centric environments.

\section{Authors}

\textbf{Core Contributors:} 
% Around the author line:
{\renewcommand{\thefootnote}{\fnsymbol{footnote}}%
Delin Qu\footnotemark[1], Qizhi Chen\footnotemark[1], Shangkun Sun\footnotemark[1], Zhaoshuo Li, Yu-Wei Chao, Xiaohui Zeng, Xuan Li, Junjie Bai, Tsung-Yi Lin, Ming-Yu Liu
\footnotetext[1]{First authors with random order}%
}

\textbf{Contributors:} Kaichun Mo, Jinwei Gu, Moo Jin Kim, Fangyin Wei, Hongchi Xia, Nic Ma

\clearpage
\bibliography{biblio}
\bibliographystyle{colm2024_conference}

\end{document}